\def\year{2018}\relax

\documentclass[letterpaper]{article} 
\usepackage{aaai18}  
\usepackage{times}  
\usepackage{helvet}  
\usepackage{courier}  
\usepackage{url}  
\usepackage{graphicx}  
\begin{document}

\title{Improbotics: Exploring the Imitation Game using \\ Machine Intelligence in Improvised Theatre}

\author{Kory W. Mathewson\textsuperscript{1,2}\ {\normalfont and} Piotr Mirowski\textsuperscript{2}\\
\textsuperscript{1}University of Alberta Edmonton, Alberta, Canada\\ \textsuperscript{2}HumanMachine, London, UK\\
korymath@gmail.com, piotr.mirowski@computer.org
}

\maketitle
\begin{abstract}
Theatrical improvisation (\emph{impro} or \emph{improv}) is a demanding form of live, collaborative performance. Improv is a humorous and playful artform built on an open-ended narrative structure which simultaneously celebrates effort and failure. It is thus an ideal test bed for the development and deployment of interactive artificial intelligence (AI)-based conversational agents, or \emph{artificial improvisors}. This case study introduces an improv show experiment featuring human actors and artificial improvisors. 
We have previously developed a deep-learning-based artificial improvisor, trained on movie subtitles, that can generate plausible, context-based, lines of dialogue suitable for theatre \cite{mathewson2017improvised}. In this work, we have employed it to control what a subset of human actors say during an improv performance. We also give human-generated lines to a different subset of performers. All lines are provided to actors with headphones and all performers are wearing headphones. 
This paper describes a \emph{Turing test}, or imitation game, taking place in a theatre, with both the audience members and the performers left to guess who is a human and who is a machine. In order to test scientific hypotheses about the perception of humans versus machines we collect anonymous feedback from volunteer performers and audience members. Our results suggest that rehearsal increases proficiency and possibility to control events in the performance. That said, consistency with real world experience is limited by the interface and the mechanisms used to perform the show. We also show that human-generated lines are shorter, more positive, and have less difficult words with more grammar and spelling mistakes than the artificial improvisor generated lines.
\end{abstract}

\begin{figure}[ht!]
    \centering
    \includegraphics[width=0.465\columnwidth]{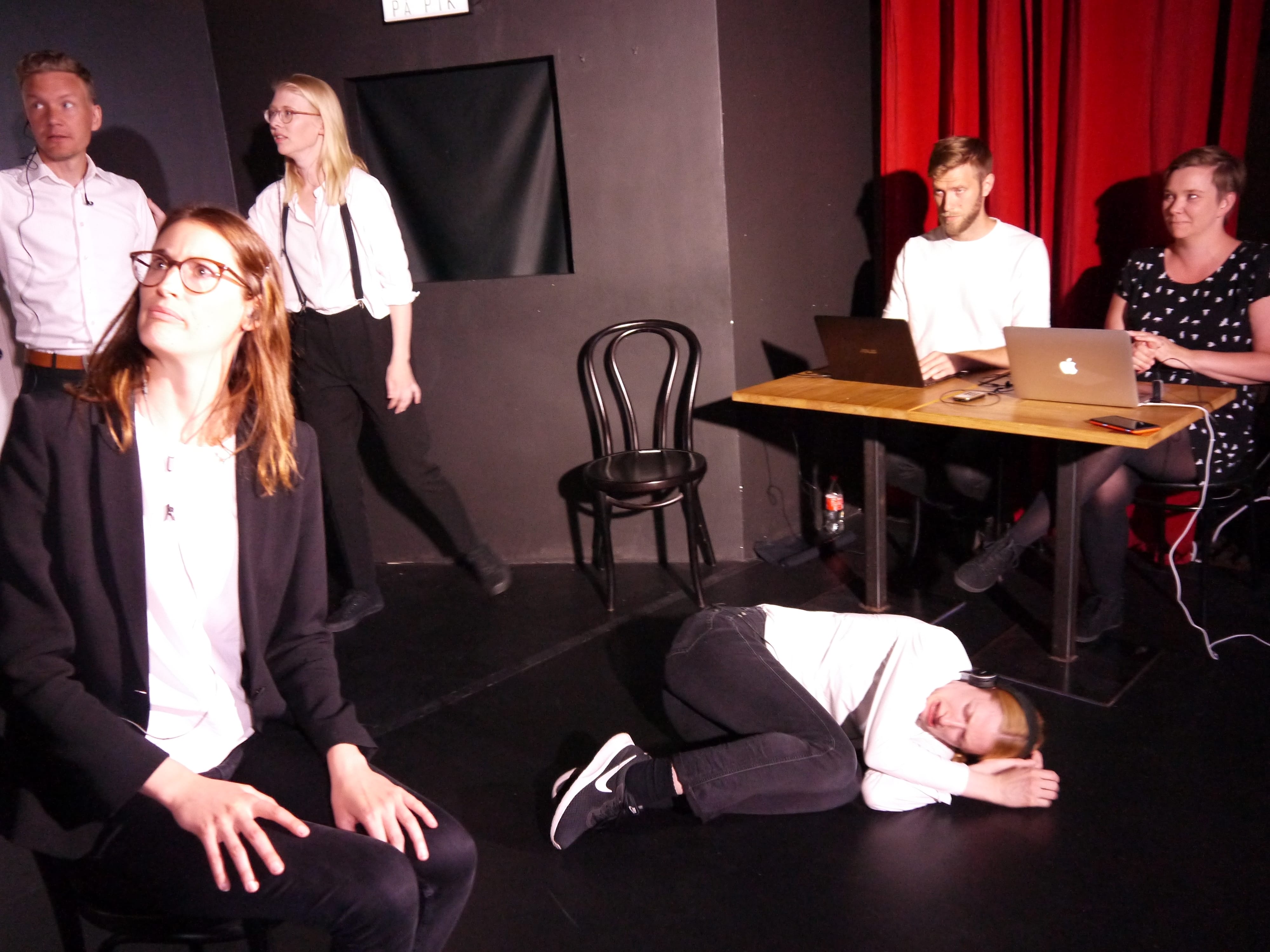}
    \includegraphics[width=0.525\columnwidth]{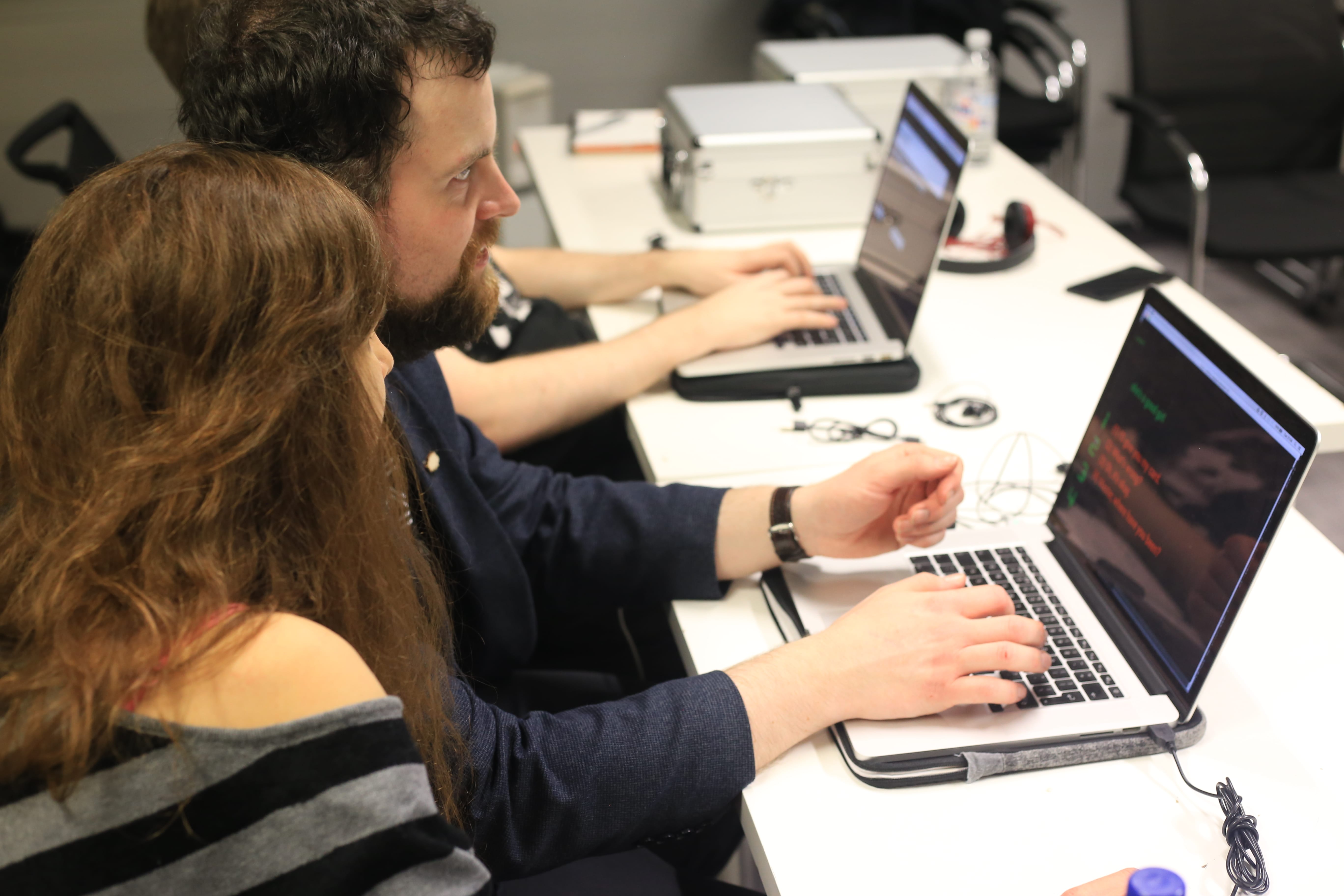}
    \caption{Illustration of two Improbotics rehearsals.}
    \label{fig:tech}
\end{figure}

\section{Introduction}
Improvisation (\emph{impro} or \emph{improv}) is a complex theatrical art form modelled on natural human interaction and demanding constant adaptation to an evolving context. It has been defined as ``real-time dynamic problem solving'' \cite{johnson2002jazz,magerko2009}. Improv requires performers to exhibit acute listening to both verbal and non-verbal suggestions coming from the other improvisors, split-second reaction, rapid empathy towards the other performers and the audience, short- and long-term memory of narrative elements, and practiced storytelling skills \cite{johnstone1979}. From an audience point of view, improvisors must express convincing raw emotions and act physically.

We agree that improvisational computational storytelling is a grand challenge in artificial intelligence (AI) as proposed by \cite{martin2016improvisational}. While success on the grand challenge might be contingent on solving open-domain conversational general artificial intelligence, there have been incremental scientific steps made progressing toward a unified system which can engage in improvised theatre in an open world \cite{zhang2018personalizing,mathewson2017improvised,guo2018improvchat,cappo2018online}. While these systems do not fully understand the interaction, they can, in spite of (or perhaps, as an improvisor would think, \emph{thanks to}) their imperfections, fuel the creativity of the performers.

\subsection{Related Work}
Research on computational improvisation often focuses on music and dance, and on how humans interact and co-create with artificial systems \cite{fiebrink2011real,hoffman2011interactive,thomaz2016computational}. Improvised theatre has also been a platform for digital storytelling and video game research for more than two decades \cite{perlin1996improv,hayes1996improvisational}. Theoreticians and practitioners have experimented with several rule- or knowledge-based methods for collaborative storytelling and digital improvisation \cite{o2011knowledge,si2005thespian,zhang2007affect,magerko2011employing}, and computer-aided interactive storytelling has been explored in video game development, aiming to create near-infinite narrative possibilities to drive longer-term player engagement \cite{riedl2006believable}. To the best of our knowledge, our case study describes the first application of deep learning-based conversational agents \cite{vinyals2015neural} to control and guide the improvised theatre performance of human actors.

Robotic performances have been explored previously \cite{breazeal2003interactive}. In 2000, Tom Sgorous performed \textit{Judy, or What is it Like to Be A Robot?} 
In 2010, the realistic humanoid robot Gemenoid F performed \textit{Sayonara}, which was later turned into a movie. Incorporating audience feedback into a robotic performance was reported by \cite{knight2011savvy}. In their work, the authors used visual sensors to track audience sentiment following a line delivered by the robotic performer, and used this information to modify the next line selection based on the feedback received. In a similar way, and as we describe in the Methods section, a human is involved in the selection of the next line produced by our conversational system.
In 2014, Carnegie Mellon University's Personal Robotics Lab collaborated with their School of Drama to produce \textit{Sure Thing} \cite{7020993}. 
In these performances, robots were precisely choreographed, deterministic, or piloted on stage \cite{hoffman2008hybrid}. These shows required the audience to suspend disbelief and embrace the mirage of autonomy. Those robot-based performances had to challenge the \textit{uncanny valley}---the idea that as the appearance of a human-like robot approaches a human likeness, human responses shift from empathy toward revulsion \cite{mori2012uncanny}. Recently, toy-like humanoid robots have been involved in improvised theatre performances \cite{magerko2009}, for instance Arthur Simone's \textit{Bot Party: Improv Comedy with Robots}\footnote{\url{http://arthursimone.com/bot-party/}} and \textit{HumanMachine: Artificial Intelligence Improvisation}\footnote{\url{https://humanmachine.live/}}. Unlike those shows, our performance does not employ robotic avatars but sends the AI-generated dialogue to a human embodiment. 

\subsection{Motivation}
Recent cinematic releases including \emph{Her} \cite{her2013} and \emph{Robot \& Frank} \cite{robotandfrank2012} explored robots interacting with humans naturally in day-to-day life; we invite live audiences to consider such interactions in a theatrical setting. We believe that theatre practitioners can embrace AI as a new tool to explore dramatic interactions and to expand the realm of stories that artists can create. This aligns with our research goal of augmenting creative abilities of humans. To test the quality of this creative augmentation, we have developed a test-bed for theatrical co-creation which places humans directly alongside machines in an improvisational performance. 

In our show \textit{Improbotics}, we explore how human performers could seamlessly perform when a machine, or another human, provides their lines. The human and machine performers work together to create a single, grounded, narrative improvisation. We combine conceptual ideas from classic improvisation and novel methods in machine learning and natural language processing. The show is inspired by improvisation game \textit{Actor's Nightmare} \cite{durang1980actor}--where one of the performers reads lines from a play and the other performers seamlessly justify these otherwise incongruous lines while progressing a narrative. This game is modified to incorporate previous work on improvised theatre alongside artificial intelligence. Specifically, this work builds on the performances of \cite{mathewson2017improvised}, \textit{HumanMachine: Artificial Intelligence Improvisation}, and Etan Muskat's \textit{Yes, Android}\footnote{\url{https://baddogtheatre.com/yes-android/}}.

This work explores wizard-of-oz style experimental methods that have been used extensively in previous human-robot interaction studies and dialogue system research \cite{riek2012wizard,edlund2008towards,fong2003collaboration,mateas1999oz}. Wizard-of-Oz style interactions with artificial intelligence controllers have been used to provide suggestions to actors into previous artistic works \footnote{\url{https://www.badnewsgame.com/}}. In these studies, humans receive inputs from an external source. The source may be another human, or the machine learning system. Importantly, the source is unknown to the human. This allows for separation between the human subjects' outputs, and the corresponding inputs. Similar to \textit{Actor's Nightmare}, the controlled humans in \emph{Improbotics} will say and justify the lines they are prescribed through emotion, intonation, and physicality. What sets this format apart from previous work is that in \textit{Improbotics} the lines depend on the context of the improvised scene. Improvisors not fed lines work to justify as the lines are not completely congruous. These justifications aim to make the scene look and feel more natural.

In a way, Improbotics can be seen as a theatrical Turing Test \cite{turing1950computing,mathewson2017turing}. Can the performers and audience discern who is delivering lines generated by a human from those delivering lines from a machine? We now cover methods to test this question.

\section{Methods}
\emph{Improbotics} is a show structure created to explore the grand challenge of artificial improvisation \cite{martin2016improvisational}. The show is composed of a cast of trained human performers (semi-professional improvisors with at least 2 years of experience). 

The cast is broken down into four roles: \emph{Cyborgs}, \emph{Puppets}, \emph{Free-will Humans}, and \emph{Controllers}.

\begin{itemize}
    \item \emph{Cyborgs} are humans who take lines via headphones from an AI-powered chatbot overseen by a \emph{CEO Controller};
    \item \emph{Puppets} take their lines via headphone from a \emph{Puppet Master Controller};
    \item \emph{Free-will Humans} are free to make up their own lines of dialogue and typically support the show's narrative; and
    \item \emph{Controllers}, of which there are two sub-roles: 1) the Puppet Master directly inputs lines for the Puppet, and 2) the CEO who inputs scene context into an AI system that generates lines of dialogue for the Cyborg. 
\end{itemize}    

\subsection{Show Structure}
\textit{Improbotics} is structured as a collection of improvised scenes. A scene starts by soliciting a suggestion for context from the audience (e.g., ``non-geographical location'' or ``advice a grandparent might give'') \cite{johnstone1979}. This provides performers with a novel context around which to situate the improvised performance, and primes the AI-system. 

The scene consists of alternating lines of dialogue, where the Free-will Humans provide dense context to the system (human or AI-based dialog model), and the Cyborg or Puppet performers respond in return. The Cyborg and Puppet performers aim to maintain the reality of the scene and to ground narratives in believable storytelling by justifying - emotionally and physically - their lines. A typical scene lasts between 3 and 6 minutes, and is concluded by the human performers when it reaches a natural ending (e.g. narrative conclusion or comical high point). The performance progresses over the course of 20-45 minutes. At the end of the show, the audience votes to guess who was a Cyborg, who was a Puppet, and who was a Free-will Human. 

Our Turing test is relatively easy to solve by an attentive audience, and similar imitation games have been documented in \cite{mathewson2017turing}. We use it instead to both draw audience engagement and to give a creative constraint to the performers, analyzing the experience of performers collaborating with interactive AI tools. Additionally, it is hard to evaluate the imitation game with a live audience because of deception  required from each attendee in a controlled but public performance setting. For this reason, we provide the Turing Test as a framework for the show though it is unlikely that audience members were tricked for the duration of the show. The audience can infer who is a Cyborg or Puppet based on typos (e.g., "We are stuck in the des\underline{s}ert?... desert!"), spelling and grammar mistakes, lack of contextual consistency, and ignored salient information or timing constraints. We discuss these points in Section \ref{sec:discussion}.

We considered a baseline show \textit{Yes, Android} that is different from \textit{Improbotics} in three aspects: 1) it relies on publicly available chatbot \textit{Cleverbot}\footnote{\url{http://www.cleverbot.com/}}, which is a general small-talk conversational agent that is trained not on movie dialogue but on user interaction, 2) there is no Master or Puppet, and 3) no guessing takes place, as the audience knows in advance who is the AI-controlled performer.

\subsection{Technical Configuration}
The technology that enables remote control of a human player consists of a laptop computer connected to an FM radio transmitter, an FM radio receiver with headphones worn by the controlled performer (Cyborg or Puppet), and a program that allows a Controller to type either the Puppet's lines, or context sent to an AI-based chatbot that will in-turn generate sentences to say by the Cyborg (see Fig.\ref{fig:tech}). We used the chatbot from HumanMachine's \emph{A.L.Ex} \cite{mathewson2017improvised,mathewson2017turing}, whose architecture is a sequence-to-sequence \cite{sutskever2014sequence} recurrent neural network \cite{hochreiter1997long} trained on movie subtitles\footnote{\url{https://www.opensubtitles.org/}} \cite{vinyals2015neural}. Full details on the model and technical configuration are excluded for brevity as they can be found in our previous work \cite{mathewson2017improvised}. Our model generates word-by-word a set of 10 candidate sentences as responses to a given input and scene context; the top 4 sentences (ranked by language model log-likelihood score) are selected and shown on the visual interface.

\begin{figure}[ht!]
    \centering
    \includegraphics[width=0.4\columnwidth]{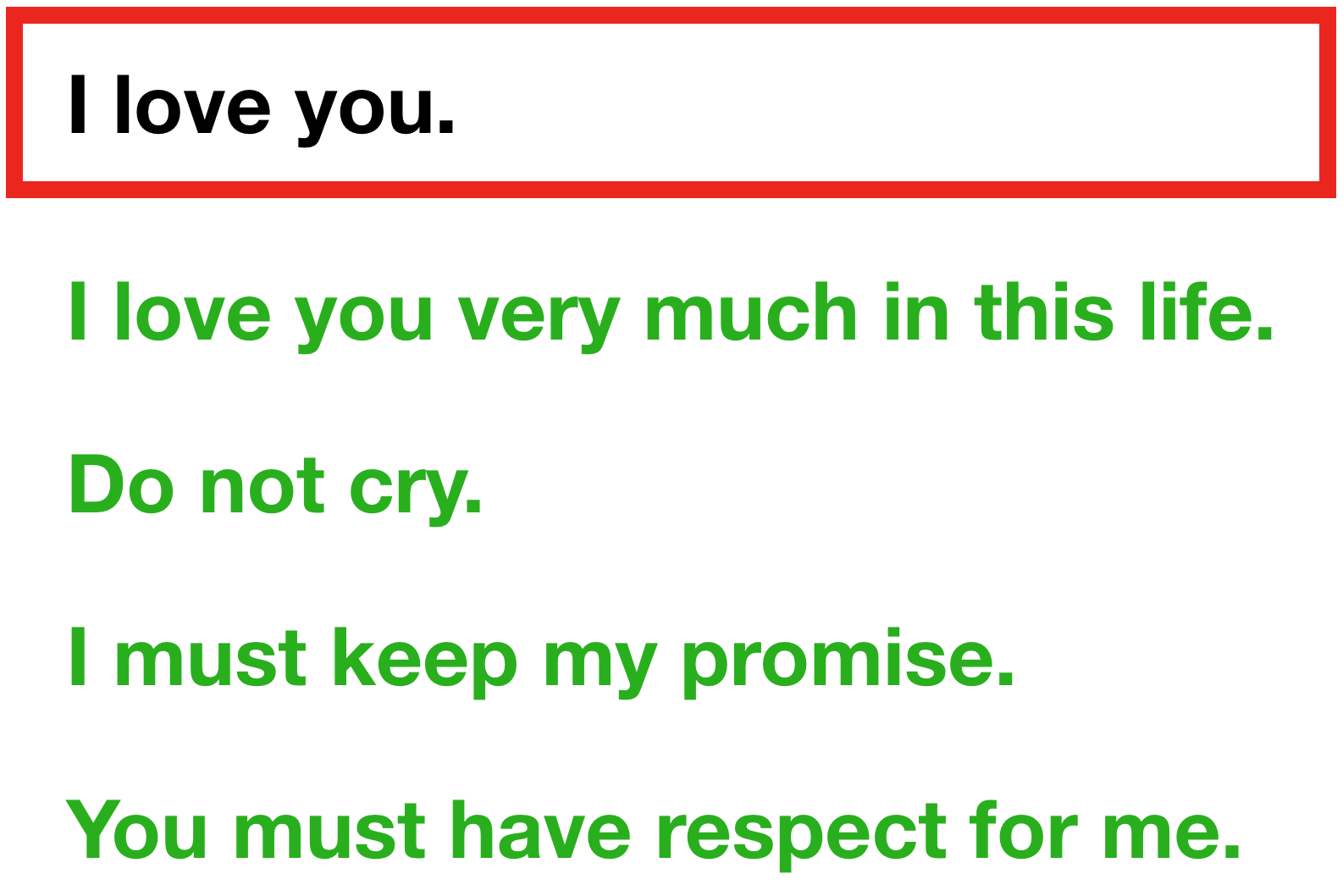}
    \caption{Simplified interface with the line of context in the red box and generated candidate responses in green.}
    \label{fig:my_label}
\end{figure}

The Puppet Master Controller interface is a webpage where the operator types lines. Once a line is typed, it is immediately converted to speech using in-browser (\textit{Google Chrome}) text-to-speech capability. The AI Controller's interface for the Cyborg is a webpage that enables to type a line of context. Once typed, that line is sent to a server running the recurrent neural network-based chatbot model, and returns up to 4 suggestions. The suggestions are displayed on the webpage, and the Controller has the choice to select as many as they want or to discard them and to type new context. When a suggestion is selected, it is immediately said via text-to-speech. As we elaborate in the Discussion section, this multiple-choice setup for the Cyborg suggestions enables the CEO to moderate what lines are sent to the Cyborg's headphones, and to curate suggestions to help drive the narrative of the performance.

All performers wear headphones, and both the Puppet and the Cyborg receive their lines through the same computer-generated voice through a \emph{text-to-speech} interface. This configuration allows for a hidden set-up where the audience, and other performers, do not know who is playing what role.

\begin{figure*}[ht!]
    \centering
    \includegraphics[width=0.75\textwidth]{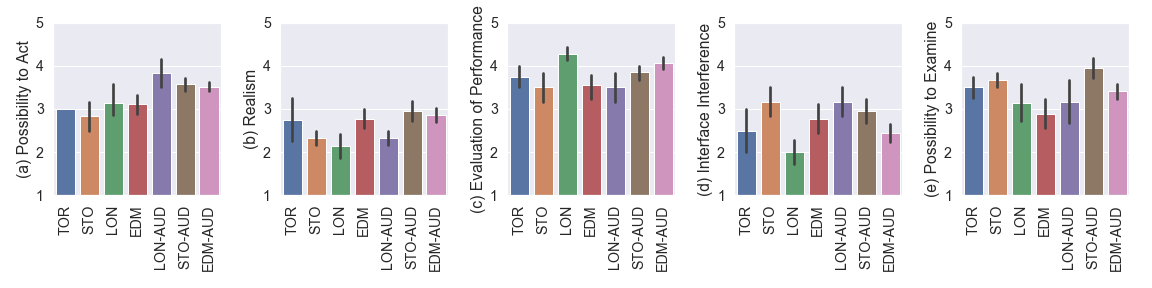}
    \caption{Audience and Performer Presence Analysis over Cities. Comparable systems were used and then analyzed by a set of performers in four cities (sample size shown in brackets): Yes, Android in Toronto (TOR, $n = 4$), Improbotics in Stockholm (STO, $n = 6$), Improbotics in London (LON, $n = 7$), and Improbotics in Edmonton (EDM, $n = 9$). Additionally, audiences were surveyed and data is presented for each city: LON-AUD ($n = 6$), STO-AUD ($n = 22$) and EDM-AUD ($n = 29$). Data presented is the average opinion over respondents in each group, within the 95 percent confidence interval.}
    \label{fig:cities}
\end{figure*}

\subsection{Evaluation}
A commonly used method of evaluating interactive performance is to address participants and audience during the show and after-the-fact, investigating experience through open questions, questionnaires or focus groups \cite{witmer1998measuring}. Our system was evaluated for \textit{humanness} based on \cite{abushawar2016usefulness}. In their work, the authors discuss that the evaluation of dialog systems should be based on comparison with interaction with real humans: this is precisely the environment we aimed to create with \textit{Improbotics}.

Post-show questionnaire questions were based on a subset of the Presence Questionnaire by \cite{witmer1998measuring}. These questions were originally developed for a small audience interacting in virtual reality domains. Questions from the original questionnaire were reworded or excluded if they pertained to immersive experiences inconsistent with  improvised theatre. The final questionnaires presented to the audience and performers measured the system on the following categories: possibility to act, realism, evaluation of performance, quality of interface, and possibility to examine the performance. In addition to quantitative survey-based evaluation, we report qualitative assessment comments. Exploring subjective opinions of performers provides us with valuable notes about human-machine co-creation. Participation in the data collection was optional. No personal identifying information was collected. Performers and audience gave informed consent, and the study was approved by the ethics review board at the University of Alberta.

\section{Results}

\begin{figure*}[ht!]
    \centering
    \includegraphics[width=0.75\textwidth]{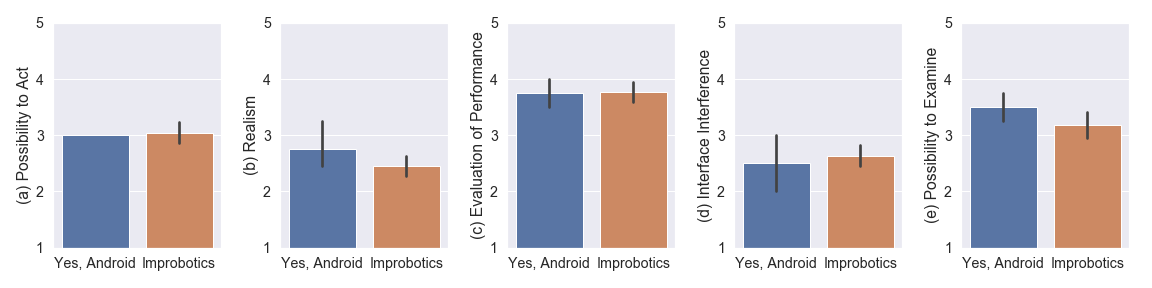}
    \caption{Performer Presence Analysis over Systems. \textit{Yes, Android} performers in Toronto ($n = 4$) used a different system than \textit{Improbotics} performers in Stockholm, London, and Edmonton ($n = 22$). This plot illustrates the comparison of analysis between the two different systems. Data presented is average opinion over respondents in group and 95 percent confidence interval.}
    \label{fig:systems}
\end{figure*}

We present here quantitative and qualitative results from experimentation with the \textit{Improbotics} system. We have deployed the experimental configuration to three locations: 1) Improbotics Stockholm, Sweden (STO, $n = 6$), 2) Improbotics London, England (LON, $n = 7$), and 3) Improbotics Edmonton, Canada (EDM, $n = 9$), where $n$ is the number of respondents. In addition to these locations, we also provide comparative results from performers in Toronto, Canada who performed in \textit{Yes, Android} (TOR, $n = 4$). We additionally present data collected from audience members who attended a show in each respective city, denoted: LON-AUD ($n = 6$), STO-AUD ($n = 22$) and EDM-AUD ($n = 29$). While audience demographic data was not collected, we infer that audiences in LON and STO were generally other improvising performers and audiences in EDM are general theatre-going patrons. Performer and audience data from multiple cities allows us to systematically measure the consistency and reproducibility of the experience on the evaluation metrics defined above.

\subsection{Quantitative Evaluation}
The questionnaire to the performers was as follows:\footnote{For the audience questionnaire, the wording of the questions was modified to reference "the performers" instead of "you".} 

\begin{enumerate}
    \item (possibility to act) How much were you able to control events in the performance?
    \item (realism) How much did your experiences with the system seem consistent with your real world experiences?
    \item (evaluation of performance) How proficient in interacting with the system did you feel at the end of the experience?
    \item (quality of interface) How much did the control devices interfere with the performance?
    \item (possibility to examine the performance) How well could you concentrate on the performance rather than on the mechanisms used to perform those tasks or activities?
\end{enumerate}

Overall, the actors were satisfied with the performance despite the limited realism of the setup (consistent between cities) and moderate interface interference. 
We note no significant difference between \textit{Improbotics} and \textit{Yes, Android}.
Improvisors from LON, who had the most rehearsals and performance opportunities with the system, rated its realism the lowest but their proficiency with it the highest, judging that the system interface did not interfere significantly with the performance. Improvisers from EDM, who had only one rehearsal, had the most trouble concentrating on the performance rather than on the interface. We infer that, with practice, the system interface interfered less with the performance and that practice increases proficiency. Audiences rated the performers as having more control of the events during the performance than the performers. 


Note that we do not split the responses from the performers of different types (Cyborg, Puppet, Puppet Master, CEO) due to the collaborative nature of improv and to the necessity to provide a single consistent show including all improvisors on the stage. Additionally, we observed that if one performer is limited in any way, it drives the entire scene down.

In addition to the qualitative questionnaire, we compare the utterance choices that the two Controllers (Puppet Master and CEO) are providing to the Puppet and the Cyborg respectively with lines from a script and lines from human improvisors. For this comparison, we selected several linguistic features (namely: syllables-per-word, words-per-sentence, proportion of difficult words, VADER sentiment \cite{vader2014hutto}, and grammatical/spelling errors) indicating the complexity of the provided sentences. 

While evaluating the quality of a dialogue interaction is difficult, these linguistic features can provide a surrogate measure of the information contained within each of the lines composing a dialogue. For the comparative lexicographical analysis we used a test set of lines from four different data sources. We analyze $L_{puppet} = 334$ lines from the Puppet Master, $L_{cyborg} = 2248$ lines generated by the dialog system, and compare with $L_{script} = 1675$ lines from two different scripts and $L_{human} = 410$ lines from Free-will Human performers in Improbotics shows. 

\begin{figure*}[ht!]
    \centering
    \includegraphics[width=0.8\textwidth]{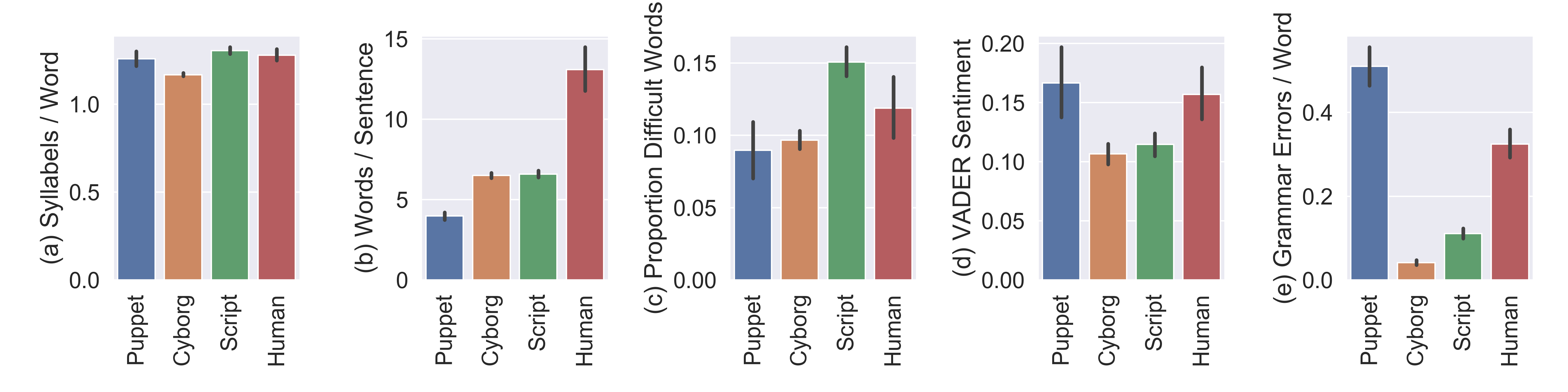}
    \caption{Comparative Lexicographical Analysis by Source. To compare the differences in text-based features we analyze a set of sentences from four different sources: 1) the Puppet Master, 2) the dialogue system or AI, 3) two published scripts ``A Streetcar Named Desire'' and ``Playing with Grown Ups'', and 4) human lines transcribed with speech recognition at a set of shows in Edmonton. Data presented is the average over each group, within the 95 percent confidence interval.}
    \label{fig:group-stats}
\end{figure*}

The scripts that we include for the analysis are similar to those used for Actor's Nightmare, namely: Tennessee Williams' ``A Streetcar Named Desire'' \cite{williams1989streetcar} and Hannah Patterson's ``Playing with Grownups'' \cite{patterson2013grownups}. As seen on Fig.\ref{fig:group-stats}, when comparing what the Master typically types to what is found in theatre scripts or what is generated by the AI, we observe that the Master/Puppet improvise with shorter lines, with considerably more grammatical or spelling mistakes (which can be explained by the time pressure on the Master to give lines to the Puppet improvisor) and with a slightly more positive VADER sentiment (likely due to the training of improvisors encouraged to ``yes, and'' by displaying positive sentiments). These results support the conclusions that human-generated lines are shorter when typed and longer when spoken. As well, human lines are more positive, have less difficult words than scripts and have more grammar and spelling mistakes than the artificial improvisor generated lines.

\subsection{Qualitative Evaluation}
In addition to the quantitative evaluation, we also asked performers to provide feedback with the following prompt: 
\textit{How would you compare performing alongside the system as compared to performing alongside a human?}

The results from this question allow us to better understand the expectations of the performers. Selected quotes from the professional improvisational performers who worked with the \textit{Improbotics} system in a variety of roles are presented below, grouped into themes.

\textbf{Theme 1: Improvising with the system is more work.}
\begin{itemize}
    \item \textit{The system throws up some real curve balls which makes it different to performing with a human.}
    \item \textit{You, as a human, have to be on your toes to validate the sometimes fun and crazy things that the Cyborg says.}
    \item \textit{The system gives more ``out-of-the-blue'' comments, and it does not feel like it is cooperating with me to make a ``perfect'' scene.}
    \item \textit{...it is a lot more work for me as a human to drive the scene, and that could be a bit lonely and cumbersome sometimes.}
\end{itemize}

\textbf{Theme 2: The system cannot tell complete stories.}
\begin{itemize}
    \item \textit{If you want to tell a story, humans tend to have to pick up the arc and carry it through, since the Cyborg rarely brings arguably important characters or plot items back.}
    \item \textit{As long as the human performers provide the improv ``platforms'' then those being controlled simply get to have fun!}
    \item \textit{I found it essential that the humans operating the system give performers enough to say; once or twice I was on stage with no lines coming through! Object work becomes super important in this instance!}
\end{itemize}

\textbf{Theme 3: Forces you to be a better improvisor.}
\begin{itemize}
    \item \textit{It makes it very important to be open and accepting. Blocking or denying of any kind only makes the ``uncanny valley'' deeper and more obvious.}
    \item \textit{...you have to be be more physical and [create] a reality which allows space for the ``curve balls'', and the cyborg's stunted performance, to make sense}
    \item \textit{...you have to listen more, and drive the scene yourself, you treat your cyborg scene partner differently--you can't rely on them completely}
\end{itemize}

\textbf{Theme 4: Like performing with a novice improvisor.}
\begin{itemize}
    \item \textit{It was like performing with a very new improvisor with strange impulses.}
    \item \textit{It takes a different mind-set, like being aware a fellow performer is very nervous and unpredictable.}
\end{itemize}

\section{Discussion}
\label{sec:discussion}
\subsection{Reflections from Professional Improvisors}
More than 20 professional improvisors have worked with the system and provided their experienced analysis and critiques which are summarized above. Their feedback largely fell into four thematic categories: 1) improvising with the system is more work, 2) the system cannot tell complete stories, 3) it forces you to be a better improvisor, and 4) it was like performing with a novice improvisor. Of these themes, two are negative (1 and 2), and two are positive (3 and 4). While working with the system is more work, this is largely due to the design of the system, to augment the humans performance. Currently, the system is focused on dialog and has no notion of a ``full story.'' Future work is needed to incorporate narrative representations into the system. The feedback that draws parallels to performing with novice improvisors is reassuring, as the goal of the system is to challenge the notion that the ``best improvisors can improvise with anyone... even a machine.''

\subsection{Deception and Problem Solving}
\textit{Improbotics} is a narrative improv show, where neither the audience, nor the improvisors, know who is a Free-will Human, who is a remotely controlled Puppet, and who is an AI-controlled Cyborg. The AI dialogue system is controlled by the CEO controller who follows the context of the scene and the narrative progression on stage, interactively producing the next line for the Cyborg performer. These lines are often nonsensical and add incongruity to the ongoing scene. The randomness of these lines was addressed directly in several of the participants' responses. While the justification of these random offers provides fun, it can also be isolating and challenging for the human performers who feel they are ``improvising with a beginner'' and need to take care of the narrative progression.

The human Puppet Master, who observes the improv scene from a hidden place, and who feeds lines to the Puppet via the earpiece, is tasked with a very difficult challenge. They need to listen to the scene and simultaneously type dialogue suitable for the next line. Alternatively, as we observed in several performances, the Puppet Master can pretend to be AI-like and through playful deception (e.g. generating more nonsensical or disconnected lines of dialogue), introduce a wild-card into the audience's mind.

We desire to push the imitation game as far as possible while creating an enjoyable performance. Thus, we encourage the improvisors to act in the most natural and intelligent way. They are expected to play to the full range of their emotions and physicality. That said, they are also aware of the conceit of the show and often they can introduce intrigue in the audience's mind by pretending to be more AI-like, more robotic. Through this ``double-bluff'' any performer can act as if they are the Puppet, or Cyborg. As anecdotal evidence, some audience members incorrectly thought that a Free-will Human was a Cyborg in two out of six Improbotics shows in London (but guessed correctly in the other ones).

If improvisation is ``real-time dynamical problem solving'' \cite{johnson2002jazz,magerko2009}, then we can see Improbotics as an optimisation problem for the performers where the main objective is producing an enjoyable theatrical performance while optimising a second meta-objective of playfully deceiving the audience.

\subsection{Lack of Contextual Consistency}
Through the comparison of the performances of the Cyborg, of the Puppet and of the classic improv game \emph{Actor's Nightmare}, we see how differently they handle two types of contextual consistencies in improvised narratives: 1) broad consistency in the general theme of the improv (e.g., domain-specific vocabulary used in the scene) and 2) fine-grained consistency in the articulation of the story (e.g., relationships between characters, character stance or world view).

In the game \emph{Actor's Nightmare}, where the improvisor reads consecutive lines for a given character, selected from a random play, those lines are typically consistent among themselves, but disconnected from the general theme of the scene. The fun of the game resides from seeing both actors striving at justifying the incongruity of juxtaposing, for instance, a classical drama with a sci-fi setting. When performing a Puppet, the performer is typically given lines from a trained human improvisor who listens to the context of the scene and types lines with both high-level thematic and fine-grained narrative consistency. Despite the best efforts of the Controller who curates the lines produced by the AI, the Cyborg typically gets inconsistent lines from the point of view of the narrative. With the topic model incorporated in our chatbot system, some thematic consistency can be maintained \cite{mathewson2017improvised}. So, the AI, when primed with words ``ship'' and ``pirate'', will likely generate sentences about sea-faring and sword-fighting. Interestingly, this is the opposite of the \emph{Actor's Nightmare}, which lacks thematic consistency. Rather than just incorporating topic in the model, future iterations of the system could include additional context. For instance, methods to re-inject scene specific content (i.e. character names, locations, relationships, noun and verb phrases) in generate responses are currently being investigated. Methods of editing prototypes and retrieving and refining candidates is an exciting area of active research \cite{2018arXiv180804776W,guu2017generating}.

\subsection{Handling Timing in Improvisation}
One of the most challenging technical hurdles for human-machine co-creation is that of timing. Verbal interaction is defined most-notably by the characteristic of rapid exchange of turns of talking. Gaps between these turns are often as short as 200ms in natural human conversation. Latencies in language processing can be on the order of 600ms \cite{levinson2015timing}. This implies that humans are often taking turns talking based on predictions of the next line of dialogue from the others in the conversation. Given this extremely short latency expectation, there is often noticeable delay for the Puppet and/or Cyborg. Our current system has a median response time of more than 2 seconds with some responses taking up to 4 seconds. The timing of these is seldom below 1 second unless we queue up additional responses to a single input and force an interruption with a potentially out-of-context follow-up line. These timing limitations are similar to \emph{Actor's Nightmare}, where one of the improvisors reads lines from a script.

Luckily, such timing latencies can be smartly hidden by proficient improvisors through emotional, nonverbal, and/or physical actions. While, in our previous work with an automated and un-curated chatbot, improvisors would typically talk over a naive robot voice responding with bad timing \cite{mathewson2017improvised}. This happened significantly less often with the Cyborg or Puppet in Improbotics, because all the people waited their turn to speak. Moreover, Cyborgs had (and used) the possibility to skip an irrelevant or outdated line.

We do however need to follow up this work with methods for better handling of timing and turn-taking, as poor timing can be a giveaway for any system imitating a human conversationalist in interactive dialogue. 

\subsection{Human Curation of Dialogue}
Currently the system is based on the natural language generation model of \cite{mathewson2017improvised}, trained on movie dialogue. We chose this corpus to train the system because it was publicly available (unlike theatre or movie scripts), because it contained informal, realistic language and because improvisors typically draw their inspiration from movies and TV series. Given that many of the movies in the source material are from over half a century ago, there are strong biases in the training material toward offensive or out-of-date references. That said, without a set of improvised dialogue transcripts, movie dialogue is the best large scale corpora available for training these models. Thus, there is a need for human moderation and curation to ensure that the system is not immediately offensive. The current system could be improved by including automated metrics for offensive language detection and removal, as presented by \cite{davidson2017automated}.

\textit{Improbotics} is focused on developing improvised dialogue in scenic improvisation. While critical to human-machine theatrical co-creation, this is only a small component of a larger automated story generation system. Incorporation of automatic plot generation techniques introduced nearly a century ago in \cite{cook1928plotto} could augment the system with directorial abilities and event-based story generation \cite{DBLP:journals/corr/MartinAHSHR17,martin2016improvisational}.

\section{Conclusion}

In this work we present \textit{Improbotics}, an improvised performance which serves as a test-bed for human-machine theatrical co-creation and can be used for improving computational dialogue-based system for live performance. The system allows for Turing test-inspired experimentation. By confronting humans to the incongruity of machines sharing the stage with them, we can both create new opportunities for comedy and explore approaches to human-machine interaction. We presented results from three geographically unique locations where the system is currently being used to perform for live audiences. We compared the \emph{Improbotics} neural network-based and movie dialogue-trained system, with the \emph{Yes, Android} baseline system, which uses an online, publicly accessible chat-bot. We presented quantitative analysis evaluating the system in five categories: realism; possibility to act; quality of interface; possibility to examine; and evaluation of performance. We present qualitative analysis from professional improvisational performers. This work focuses on improv, but this research can be applied to other areas of human-machine physical and verbal interaction.

\section{Acknowledgements}

The authors wish to thank Patrick Pilarski, Lana Cuthbertson, Alessia Pannese, Kyle Mathewson and Julian Faid for helpful discussion. We extend gratitude to the cast members of Improbotics (Leif Andreas, Jill Bernard, Paul Blinov, Sarah Castell, P\"ar Dahlman, Sarah Davies, Kerri Donaldson, Julia Eckhoff, Jenny Elfving, Roel Fos, Michael H\.akansson, Lina Hansson, Mino Ingebrigtsen, Maria Lindberg Reinius, Arfie Mansfield, Adam Meggido, Katherine Murray-Clarke, Chase Padgett, Shama Rahman, Tommy Rydling, Preeti Singh, Linn Sparrenborg, Linus Tingdahl, Harry Turnbull, James Whittaker), Yes, Android (Etan Muskat and others) and Rapid Fire Theatre for their participation in this study. Finally, we thank the anonymous reviewers for their thoughtful comments and feedback.


\bibliographystyle{aaai}
\bibliography{refs}

\end{document}